\documentclass[letterpaper, 10 pt, conference]{format/ieeeconf}   

\IEEEoverridecommandlockouts                              
\usepackage{graphicx}
\usepackage[table,dvipsnames]{xcolor} 
\usepackage{amsmath,amssymb,amsfonts}
\usepackage{mathtools}                
\usepackage{algorithm}
\usepackage[noend]{algpseudocode}     
\usepackage{array,booktabs, multirow, eqparbox}
\usepackage[caption=false,font=footnotesize]{subfig}
\usepackage{tikz}
\usepackage{url,csquotes,pifont,tablefootnote,textcomp,scalerel}
\usepackage{siunitx}
\usepackage{balance}
\usepackage{cite}

\graphicspath{ {./figures/} }

\newcommand\copyrighttext{%
  \scriptsize\centering
  \textcopyright\ 2026 IEEE. Personal use of this material is permitted.
  Permission from IEEE must be obtained for all other uses, in any current or future
  media, including reprinting/republishing this material for advertising or promotional
  purposes, creating new collective works, for resale or redistribution to servers or
  lists, or reuse of any copyrighted component of this work in other works.\\
  Accepted manuscript. Accepted for publication in the 2026 IEEE National Aerospace
  and Electronics Conference (NAECON). DOI not yet assigned.}
\newcommand\copyrightnotice{%
\enlargethispage{-0.7in}%
\begin{tikzpicture}[remember picture,overlay]
\node[anchor=south,yshift=10pt] at (current page.south) {\fbox{\parbox{\dimexpr\textwidth-\fboxsep-\fboxrule\relax}{\copyrighttext}}};
\end{tikzpicture}%
}

\newcommand{\shrteq}{\mathrel{\scalebox{0.73}[0.9]{$=$}}}
\title{\LARGE \bf
Loco-Loco-RL: \underline{Lo}w-\underline{Co}st Terrain Mapping for Humanoid \underline{Loco}motion with \underline{R}einforcement \underline{L}earning 
}

\author{Jordan Dowdy$^{1^{*}}$, Gryffin Reizian$^{2}$, and Jean Chagas Vaz$^{2}$
\thanks{$^{1^{*}}$J. Dowdy, a PhD student, and $^{2}$G. Reizian, an undergraduate student, are both with the Department of Electrical and Computer Engineering at the University of Louisville.~{\tt\small jordan.dowdy@louisville.edu}, {\tt\small gryffin.reizian@louisville.edu}}
\thanks{$^{2}$Dr. Jean Chagas Vaz is with the Faculty of Electrical and Computer Engineering at the University of Louisville, Louisville, KY 40208, USA. {\tt\small jean.chagasvaz@louisville.edu}}
\thanks{ $^{*}$direct all correspondence to this author.}
}

\begin{document}

\maketitle
\thispagestyle{empty}
\pagestyle{empty}
\copyrightnotice

\begin{abstract}
Informative terrain perception is important for robust reinforcement learning policies in humanoid locomotion. Still, common sensors such as depth cameras and LiDARs incur high cost, power, and processing overhead while often producing redundant, high-resolution data. This work uses a low-cost time-of-flight sensor to provide a compact $3D$ local terrain representation for humanoid locomotion. To efficiently use this sparse exteroceptive input, we introduce a token-compressed temporal transformer policy. Proprioceptive and terrain observations are tokenized and processed by a self-attention multi-head transformer to capture within-timestep relationships between observation terms. The attended tokens are then compressed through an MLP-based latent-space token compression module before being stored in a rolling $15$-timestep history. A second cross-attention multi-head transformer extracts temporal locomotion features from this compact history for policy learning. By compressing tokens before temporal aggregation, the architecture preserves important terrain-observation structure while limiting the dimensional growth of attention over observation histories. We validate our method through sim-to-real transfer on physical hardware using a terrain-based locomotion benchmark, demonstrating robust humanoid terrain walking with low-cost local terrain sensing.
 \end{abstract}

\section{INTRODUCTION}

Humanoid locomotion in dynamic, unstructured terrain remains a challenging problem due to the coupled requirements of perception, whole-body coordination, and dynamic balance. Reinforcement learning (RL) has emerged as a powerful framework for humanoid control as it can learn coordinated behaviors directly from interaction. Still, the robustness of learned policies is often strongly tied to the quality and form of the observation space. In particular, terrain perception is frequently a limiting factor for successful deployment outside of highly structured environments.
Most terrain-aware locomotion systems use dense exteroceptive sensing, typically using depth cameras or LiDAR, to provide terrain information. While these sensors provide rich spatial information, they often produce substantially more raw data than needed, which is later compressed or downsampled before being consumed by the Policy. They also introduce practical drawbacks for humanoid platforms. High-resolution sensing increases computational and memory requirements, adds power and integration overhead. 

\begin{figure}[!t]
	\centering
	\vspace{1.5mm}
	\includegraphics[width= 3.4in]{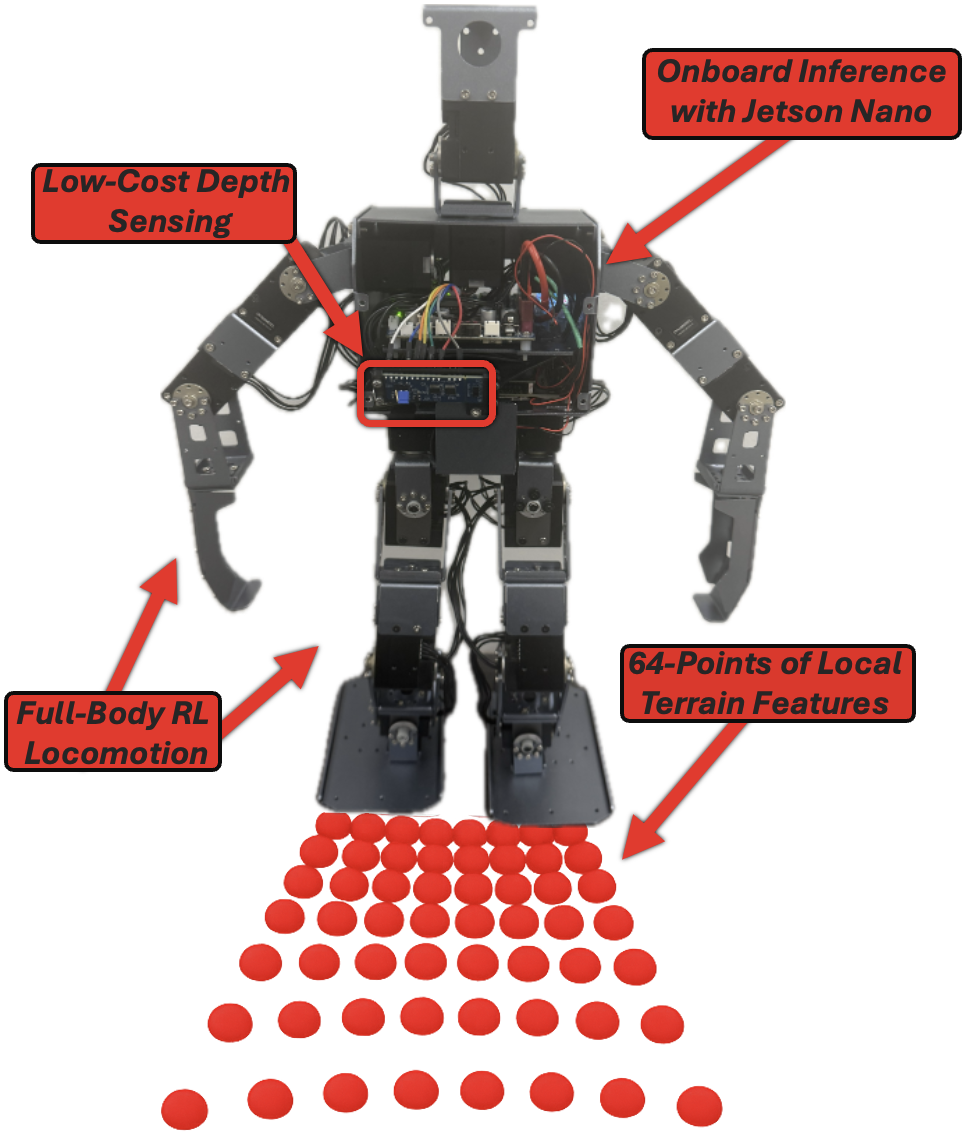}
	\caption{An overview of our incorporation of a multi-array Time of Flight (ToF) sensor to capture a local area of terrain-based features in front of the robot, and incorporate them into a full-body RL locomotion policy.}
    \vspace{-4mm}
	\label{fig_Conceptual_Pic}
\end{figure}

This observation motivated the use of a low-cost exteroceptive sensor for humanoid locomotion. In this work, we investigate whether a small time-of-flight (ToF) sensor with only $64$ spatial points and a limited field of view can provide sufficient terrain information for robust RL-based locomotion on a humanoid robot. Although such a sensor has a lower resolution than conventional depth cameras or LiDAR, it offers attractive deployment characteristics, including low cost, minimal processing requirements, reduced power consumption, and compact onboard integration. The central question is whether a policy can learn to extract and exploit the informative terrain structure contained in such sparse measurements.

To address this, we design a token-compressed temporal transformer policy. Each deployable observation term, including proprioceptive measurements, previous actions, velocity commands, and the sparse height scan, is mapped into a learned token representation. A self-attention multi-head transformer first fuses the current-timestep tokens, allowing the Policy to model relationships between body-state and terrain observations. The attended token set is then compressed into a small number of lower-dimensional latent tokens, which are then inserted into a rolling temporal history. A second cross-attention multi-head transformer queries this compact history to extract locomotion-relevant temporal features for the actor policy. Rather than relying on dense perception or directly attending over long high-dimensional observation histories, our approach learns a compact latent representation that preserves task-relevant terrain structure while reducing sensing and computational overhead.

\subsection*{Paper Contributions}

The paper's main contributions are:
\begin{enumerate}
    \item Use of a low-resolution ToF sensor for $3D$ terrain representation.  
    \item A tokenized observation representation that combines proprioceptive measurements and sparse local terrain samples.
    \item A token-compressed transformer policy that compresses current-timestep attention tokens before temporal aggregation with cross-attention over a rolling history.
    \item Sim-to-Real transfer on the Robotis OP3 humanoid with a terrain-based locomotion benchmark.
\end{enumerate}

\subsection*{Article Structure}
This paper contains six sections: Section II reviews the literature on humanoid locomotion, including both classical and reinforcement learning approaches, as well as RL approaches that use terrain-based features. Section III provides a formal theoretical definition of our work and details our reinforcement learning framework. Section IV discusses the approach to validating our work through experimentation, and Section V presents and discusses the results of our experiments. Finally, Section VI provides an overview of the presented work, including a discussion of the problems encountered and future improvements.

\section{RELATED WORK}
\subsubsection{Classical Humanoid Locomotion}
Traditional Model Predictive Control (MPC) of a robot's Zero Moment Point (ZMP) is straightforward when mapping most bipedal walking robots, utilizing Linear Inverted Pendulum and Cart-Table methodology, and allowing for very fine control over a bipedal robot's motion~\cite{10.1007/978-981-95-1103-7_24,HALDAR2023122}. Limitations arise when facing ``flexible joints and linkages,'' but these problems are solvable with more flexible MPC formulations~\cite{biomimetics10010030}.

\subsubsection{Reinforcement Learning Humanoid Locomotion}
Reinforcement learning, the use of rewards to fine-tune an algorithm to perform specific tasks, offers a unique advantage for humanoid locomotion in robotics over MPC due to the reduced need for hand-designed trajectories and online optimization. It can be integrated with other control methods~\cite{JDHumanoids25}, can handle extreme environments it is trained for~\cite{doi:10.1126/scirobotics.adi9579}, and can be trained only on visual data~\cite{8778209}. More recent work on humanoid locomotion has also shown that policies can use histories of proprioceptive observations and actions to adapt across challenging terrain, including transformer-based policies~\cite{radosavovic2024challengingterrain}.

\subsubsection{Transformer-Based Policy Architectures}
Transformers introduced a self-attention architecture that processes tokenized inputs without relying solely on recurrent or convolutional structures~\cite{vaswani2017attention}. In reinforcement learning, transformer variants have been used as memory models for partially observed tasks, where attention over previous observations can improve temporal reasoning compared with a fixed recurrent state~\cite{parisotto2020stabilizing}. These works motivate policy architectures that first tokenize heterogeneous observations and then use attention to fuse same-timestep terms and temporal history.

\subsubsection{Terrain-Based Features in Reinforcement Learning Locomotion}
The ability for a humanoid robot to see is an important facet of the senses it requires, but depth helps map terrain in more detail, for example, using time-of-flight sensors. The ability for a robot to gauge depth proves an invaluable tool~\cite{sun2025dpldepthonlyperceptivehumanoid}, especially when moving in uneven terrain due to the ability to predict or model said terrain~\cite{9891830}. Prior legged locomotion work has further shown the value of combining exteroceptive and proprioceptive inputs for terrain-aware control~\cite{miki2022learning}, while recent humanoid parkour methods use vision-based whole-body policies to handle larger terrain changes and obstacles~\cite{zhuang2024humanoidparkour}.

\section{METHODOLOGY}
This section outlines the system's hardware and architecture, along with a formal definition of the token-compressed temporal transformer policy. Implementation details for RL training are also provided, including policy rewards, the simulation environment, domain randomization, policy observations, and other RL training parameters.

\subsection{Reinforcement Learning Policy}

\subsubsection{Formal Definition}
For an RL policy modeled as a Markov Decision Process (MDP), the agent's goal is to maximize a reward function $G_t$, consisting of the current state's $\mathcal{S}$, possible future actions $\mathcal{A}$, the agent's reward $\mathcal{R}$ for a given action in $\mathcal{A}$, and a discount factor $\gamma^k$.   
\begin{equation}
    G_t = \sum^{\infty}_{k=0} \gamma^k \mathcal{R}(\mathcal{S}_{t+k},\mathcal{A}_{t+k}),\label{eq:CDR}
\end{equation}

Where a policy $\pi(\mathcal{A}|\mathcal{S})$ relates the states of the agent to possible actions. Where an optimal policy $\pi^*$ maximizes the functions,
\begin{equation}
    \pi^* \shrteq \arg \max_{\pi}\left[ \mathbb{E}_{\pi} \sum^{\infty}_{k=0} \gamma^k \mathcal{R}(\mathcal{S}_{t},\mathcal{A}_{t}|\mathcal{S}_0 \shrteq \mathcal{S}) \right],
    \label{eq:Vpi}
\end{equation}
\begin{equation}
    \pi^* \shrteq \arg \max_{\pi} \left[ \mathbb{E}_{\pi} \sum^{\infty}_{k=0} \gamma^k \mathcal{R}(\mathcal{S}_{t},\mathcal{A}_{t}|\mathcal{S}_0\shrteq\mathcal{S},\mathcal{A}_0\shrteq\mathcal{A}) \right].
    \label{eq:Gpi}
\end{equation}

Where $\pi^*$ must maximize both the expected reward for a given state $\mathcal{S}$ as seen in Eq.\eqref{eq:Vpi}, and the reward for a given state $\mathcal{S}$ taking an action $\mathcal{A}$ shown in Eq.\eqref{eq:Gpi}.

\subsubsection{Token-Compressed Temporal Transformer Policy}
Let $o_t \in \mathbb{R}^{n_o}$ denote the deployable proprioceptive observations at time $t$, let $D_t\in\mathbb{R}^{n_D}$ denote the ToF terrain observation, and let $p_t\in\mathbb{R}^{n_p}$ denote privileged signals available only during training. The deployable actor input,
\begin{equation}
    \bar{o}_t = [o_t;\,D_t],
\end{equation}
which is represented internally as $K_a=11$ semantic observation terms $\{x_t^k\}_{k=1}^{K_a}$. The critic receives the privileged input
\begin{equation}
    \tilde{o}_t = [o_t;\,D_t;\,p_t],
\end{equation}
represented as $K_c=19$ terms. For each actor term, a learned MLP tokenizer maps the input to a token
\begin{equation}
    h_t^k = T_k(x_t^k;\eta_k) + e_k,
    \quad h_t^k\in\mathbb{R}^{d_h},
\end{equation}
where $e_k$ is a learned term-identification embedding and $d_h=128$. The actor token set is
\begin{equation}
    H_t^a = [h_t^1,\ldots,h_t^{K_a}] \in \mathbb{R}^{K_a\times d_h}.
\end{equation}
Current-timestep token fusion is performed with a self-attention multi-head transformer,
\begin{equation}
    \hat{H}_t^a = F_{\mathrm{self}}(H_t^a;\theta_{\mathrm{self}}),
\end{equation}
which models relationships between observation terms within the same timestep. The attended token set is then flattened and compressed by an MLP into $N_z=4$ latent tokens with dimension $d_z=64$:
\begin{equation}
    Z_t^a = \operatorname{reshape}_{N_z\times d_z}
    \left(C_{\psi}(\operatorname{vec}(\hat{H}_t^a))\right) + E_z,
\end{equation}
where $C_{\psi}:\mathbb{R}^{K_a d_h}\rightarrow\mathbb{R}^{N_z d_z}$ and $E_z\in\mathbb{R}^{N_z\times d_z}$ is a learned slot embedding. This compression step limits the dimensional growth that would otherwise occur when storing high-dimensional token sets across time.

Temporal memory is maintained as a rolling history of $L=15$ timesteps. After inserting the current compressed tokens, the chronological memory is
\begin{equation}
    M_t^a = [Z_{t-L+1}^a;\ldots;Z_t^a] + E_{\tau}
    \in \mathbb{R}^{(L N_z)\times d_z},
\end{equation}
where $E_{\tau}$ denotes learned time embeddings added to the stored tokens. A cross-attention multi-head transformer uses the current compressed tokens as queries and the temporal memory as keys and values:
\begin{equation}
    \bar{Z}_t^a = F_{\mathrm{cross}}(Q=Z_t^a,\,K=M_t^a,\,V=M_t^a;\theta_{\mathrm{cross}}).
\end{equation}
A learned summary query maps the temporal output tokens to a single context vector $z_t^{\mathrm{temp}}\in\mathbb{R}^{d_z}$. A second summary of the current compressed tokens is added as a residual,
\begin{equation}
    z_t = z_t^{\mathrm{temp}} + z_t^{\mathrm{cur}},
\end{equation}
where $z_t\in\mathbb{R}^{d_z}$ is the feature used by the actor head. The Policy predicts the mean action vector
\begin{equation}
    \mu_t = \pi_{\theta}(z_t)\in\mathbb{R}^{18},
\end{equation}
where the $18$ actions correspond to $12$ leg joints followed by $6$ arm joints. During training, the critic follows the same token-compressed temporal structure with the additional privileged terms in $\tilde{o}_t$ for value estimation. These privileged observations are not required during deployment.

\subsection{Hardware, Simulation Environment, and Training}
\begin{figure}[!t]
    \vspace{1.5mm}
    \centering
    \setlength{\fboxsep}{-0.0pt}%
    \setlength{\fboxrule}{2.0pt}%
    \fbox{%
        \includegraphics[width=0.98\linewidth]{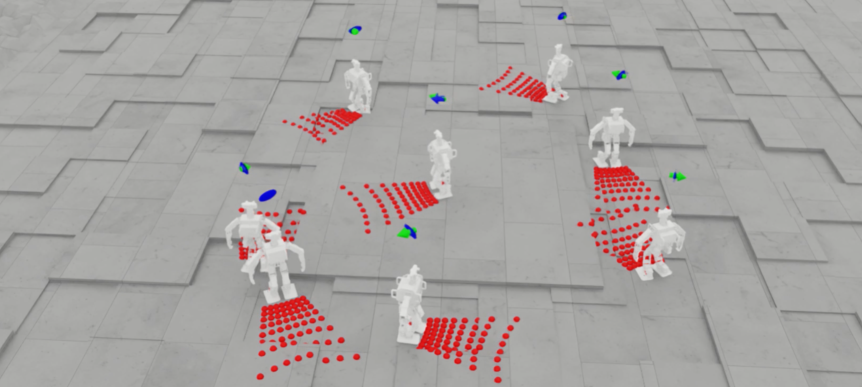}%
    }

    \caption{Policy with the Robotis OP3 and local terrain depth sensing inside of Isaac Sim.}
    \label{fig:op3Env}

\end{figure}
The local terrain sensor is based on an ST VL53L8-family multi-zone ToF module, which provides an $8 \times 8$ depth grid for local terrain observation. For policy training and deployment, the Robotis OP3 humanoid was selected. For physics simulation, Isaac Sim~\cite{NVIDIA_Isaac_Sim} and its reinforcement learning framework, Isaac Lab~\cite{IsaacGym, IsaacLab} were used. An example of the simulation environment is shown in Fig.~\ref{fig:op3Env}. An Nvidia RTX $5090$ GPU trained the Policy, running $4096$ parallel environments, with a physics solver frequency of $500.0$Hz and policy control frequency of $50.0$Hz, taking approximately $\sim80$ hours to complete.

The trained Policy uses the token-compressed temporal transformer described above. The $8\times8$ ToF height scan is treated as one $64$-value observation term and tokenized with an MLP rather than processed by a Convolutional Neural Network (CNN). Compatible observation terms share tokenizer groups, and each term is mapped to a $128$-dimensional token with an added learned term-identification embedding. The actor uses $11$ current-timestep tokens, while the critic uses $19$ tokens due to additional privileged observations. One self-attention transformer block fuses the current tokens, after which an MLP compresses the attended token set into four $64$-dimensional latent tokens. These compressed tokens are stored in a $15$-timestep rolling history and processed with a cross-attention transformer, where the current compressed tokens query the temporal memory. The resulting $64$-dimensional temporal feature is passed to a three-layer actor head that outputs the $18$-dimensional mean action vector.

For hardware deployment, the ToF sensor is interfaced with an ESP32 microcontroller, which streams local terrain-depth measurements to an onboard Jetson Nano. A Robot Operating System (ROS) serial node collects the streamed ESP32 data and formats it to an $8 \times 8$ terrain scan before being published to the policy inference node. Policy inference is performed locally on the integrated Jetson Nano mounted inside the Robotis OP3.

\subsubsection{Observations, Actions, and Rewards}
Policy observations consist of eight deployable semantic groups, which are internally split into $11$ actor terms for tokenization. The critic receives eight additional privileged training-only groups, resulting in $19$ critic tokens and providing a richer value-function estimate during training. The system's observations are detailed in Table~\ref{tab:obsTerms}, which lists the relevant observation groups for the actor and the critic. Key stability features, such as the center of mass, centroidal angular momentum, and capture point, are available only to the critic during training and are not required for deployment. This keeps the deployed actor restricted to onboard measurements while allowing the critic to use a richer description of the training-time state. 

The produced Policy predicts changes in $18$ joint positions from a nominal pose, with $12$ joints in the legs and $6$ in the arms. Action scalers of $0.15$ and $0.25$ are applied to the policy outputs for the legs and arms, respectively.

\begin{table}\centering
\vspace{2mm}
\caption{Observation Terms and Privileged Access}\label{tab:obsTerms}
\setlength{\tabcolsep}{4pt}
\renewcommand{\arraystretch}{1.35}
\begin{tabular}{llcc}
\toprule
\textbf{Observation Term} & \textbf{Symbol} & \textbf{Actor} & \textbf{Critic} \\
\midrule
\multicolumn{4}{l}{\textit{Available at deployment}}\\
\midrule
Linear Acceleration & $\mathbf{a}_{b} \in \mathbb{R}^{3}$ & \checkmark & \checkmark \\
Base Angular Velocity & $\boldsymbol{\omega}_{b} \in \mathbb{R}^{3}$ & \checkmark & \checkmark \\
Projected Gravity & $\mathbf{g}_{p} \in \mathbb{R}^{3}$ & \checkmark & \checkmark \\
Joint Positions & $\mathbf{q} \in \mathbb{R}^{18}$ & \checkmark & \checkmark \\
Joint Velocities & $\dot{\mathbf{q}} \in \mathbb{R}^{18}$ & \checkmark & \checkmark \\
Velocity Commands & $\mathbf{u}_{\mathrm{cmd}} \in \mathbb{R}^{3}$ & \checkmark & \checkmark \\
Previous Actions & $\mathbf{a}_{t-1} \in \mathbb{R}^{18}$ & \checkmark & \checkmark \\
Terrain Depth Map & $\mathbf{D}_{ToF} \in \mathbb{R}^{8\times8}$ & \checkmark & \checkmark \\
\midrule
\multicolumn{4}{l}{\textit{Privileged (training only)}}\\
\midrule
Linear Velocity & $\mathbf{v}_{b} \in \mathbb{R}^{3}$ & -- & \checkmark \\
Center of Mass & $\mathbf{c}_{b} \in \mathbb{R}^{3}$ & -- & \checkmark \\
Centroidal Angular Momentum & $\mathbf{L}_{b} \in \mathbb{R}^{3}$ & -- & \checkmark \\
Foot Contact & $\mathbf{c}_{f} \in \mathbb{R}^{2}$ & -- & \checkmark \\
Foot Air-Time & $\mathbf{t}_{\mathrm{air}} \in \mathbb{R}^{2}$ & -- & \checkmark \\
Foot Height & $\mathbf{h}_{f} \in \mathbb{R}^{2}$ & -- & \checkmark \\
Foot Forces & $\mathbf{F}_{c} \in \mathbb{R}^{2\times 3}$ & -- & \checkmark \\
Capture Point & $\mathbf{p}_{f} \in \mathbb{R}^{2\times 3}$ & -- & \checkmark \\
\bottomrule
\end{tabular}
\vspace{-1mm}
\end{table}

The rewards for our Policy follow a similar approach to our previous works~\cite{JDHumanoids25,JDiros26,JDcase25,JDsii25}, where a more formal description and explanation of these rewards are provided. The rewards used for this work policy are listed in Table~\ref{tab:rwdTerms}, including each reward, its general expression, and its associated weight.

\begin{table}
\vspace{2.25mm}
\centering
\caption{Reward and Penalty Terms}\label{tab:rwdTerms}
\setlength{\tabcolsep}{3pt}
\renewcommand{\arraystretch}{1.35}
\begin{tabular}{lccl}
\toprule
\textbf{Reward Term}                    & \textbf{General Expression}                                      & \textbf{Weight}       \\
\midrule
Feet Air Time                & $\sum^2_{i=1}(t_{\mathit{air},i} - 0.3)$                              &5.0           \\
Linear Velocity Error        & $\mathit{exp}(-(||\upsilon_{x,y}^{\mathit{des}} - \upsilon_{x,y}^{\mathit{cur}}||/0.25) )$                    &7.0          \\
Angular Velocity Error       &$\mathit{exp}(-(||\omega_{\mathit{z}}^{\mathit{des}} - \omega_{\mathit{z}}^{\mathit{cur}}||/0.25) )$                    &5.0           \\
Gait                         & $\prod^2_{i=1}\ell_{\mathit{sync},i}\cdot\prod^2_{i=1}\ell_{\mathit{async},i}$                     &10.0  \\
\midrule
\textbf{Penalty Term}                   & \textbf{General Expression}                                           & \textbf{Weight}       \\
\midrule
Action Smoothness                & $||\mathcal{A}_t - \mathcal{A}_{t\text{-}\scaleto{1\mathstrut}{5.5pt}}||$                  &-0.6         \\
Action Magnitude                & $||\mathcal{A}_t||$                  &-0.175         \\
Base Motion                      & $| \omega_x \text{+} \omega_y |$      &-1.0         \\
Base Orientation                 & $||\textbf{G}_{x,y}||$                           &-1.0           \\
Foot Slippage                    & $||\textbf{C} \cdot \upsilon_{\mathit{foot}}||$                         &-0.5           \\
Joint Position Delta              & $\sum^{18}_{h=1}||\theta_{\mathit{i}}^{nom.}-\theta_i^{act.}||$                           &-0.6           \\

Joint Torque                     & $||\tau||$                          &-$5.0\cdot10^{\text{-}4}$           \\
Joint Acceleration               & $||{\Ddot{\theta}}||$                           &-$2.5\cdot10^{\text{-}5}$           \\
Joint Velocity                   & $||\dot{\theta}||$                           &-$2.5\cdot10^{\text{-}3}$         \\
Thigh Contact                          & $\sum\textbf{C}_{thigh}$  &-1.5 \\
Knee Contact                          & $\sum\textbf{C}_{knee}$ &-1.0 \\
Ankle Contact                          & $\sum\textbf{C}_{ankle}$ &-1.0 \\
Joint Position Limits                  & $\theta >\theta_{max}$ &-1.0 \\
\bottomrule
\end{tabular}
\vspace{-1mm}
\end{table}

\subsection{Sim-to-Real}
Domain randomization is used during policy training by modifying different properties of the robot. This work uses custom domain randomization terms in Isaac Lab, allowing for joint maximum torque and joint maximum velocity to be randomly set, along with pre-existing modifiers such as joint friction, armature, link mass, and link center of mass. Joint-related properties are changed to be $\pm50.0\%$ from their initialized quantities, eliminating the need for a pre-world-to-sim actuator modeling before training. The domain randomization quantities used are shown in detail in Table~\ref{tab:DomRand}. Additionally, a terrain-based curriculum is used with multiple rough, boxed, sloped, and custom uneven terrain types. Actuator commands are additionally delayed by $(0.0,0.02)$ seconds, with the max delay equal to a policy control cycle.

\begin{table}\centering
\caption{Domain Randomization}\label{tab:DomRand}
\setlength{\tabcolsep}{4pt}
\renewcommand{\arraystretch}{1.5}
\begin{tabular}{lccl}
\toprule
\textbf{Term}              & \textbf{Operation} & \textbf{Range}\\ 
\midrule
Foot Friction  &  New    & $(0.4, 1.1)$
\\
Center of Mass    & Add   & $(\text{-}0.025,0.025)$                    \\
Link Mass  & Scale        & $(0.8,1.3)$         \\
Body Velocity & Add & $(\text{-}\hspace{0.1em}0.25, 0.25)$ \\
Body Position & Add & $(\text{-}\hspace{0.1em}0.5,0.5)$ \\
Body Orientation & Add & $(\text{-}\hspace{0.1em}0.02,0.02)$ \\
Initial Joint Pos.  & Scale  & $(\text{-}\hspace{0.1em}0.3, 0.3)$ \\
Initial Joint Vel.  & Scale  & $(\text{-}\hspace{0.1em}1.5, 1.5)$ \\
Joint Armature  & Scale  & $(\hspace{0.1em}0.5, 1.5)$ \\
Joint Friction  & Scale  & $(\hspace{0.1em}0.5, 1.5)$ \\
Joint Max Torque & Scale & $(\hspace{0.1em}0.5, 1.5)$ \\
Joint Max Velocity & Scale & $(\hspace{0.1em}0.5, 1.5)$ \\
\bottomrule
\end{tabular}    
\end{table}

\section{EXPERIMENTATION AND VALIDATION}
This section aims to validate our proposed humanoid locomotion policy through real-world deployment and experimentation. The experiments performed on our Policy assess velocity-tracking performance and rough-terrain capabilities through a multi-terrain locomotion experiment.

\subsection{Velocity Tracking Performance}
Velocity-tracking performance is evaluated on flat terrain using a pre-determined sinusoidal signal for all command velocities. A total of $10$ trials were performed on hardware, with the sinusoidal command having a frequency of $0.25$ Hz and an amplitude of ~$0.75$ m/s or $1.0$ rad/s for linear and angular velocities, respectively. 

\subsection{Multi-Terrain Locomotion Validation}
Policy terrain locomotion validation was performed on rough/rocky terrain to evaluate whether the local ToF height scan provides enough information for terrain-aware walking. This experiment tests how well the Policy uses sparse local terrain features to preserve stability while following velocity commands on uneven ground.

\section{RESULTS AND DISCUSSION}
This section presents the data collected during experimentation and performance measurement, highlighting both the strengths and limitations of the Policy.

\subsection{Velocity Tracking Metrics}
\begin{figure}
    \vspace{1.50mm}
    \centering
    \includegraphics[width=0.98\linewidth]{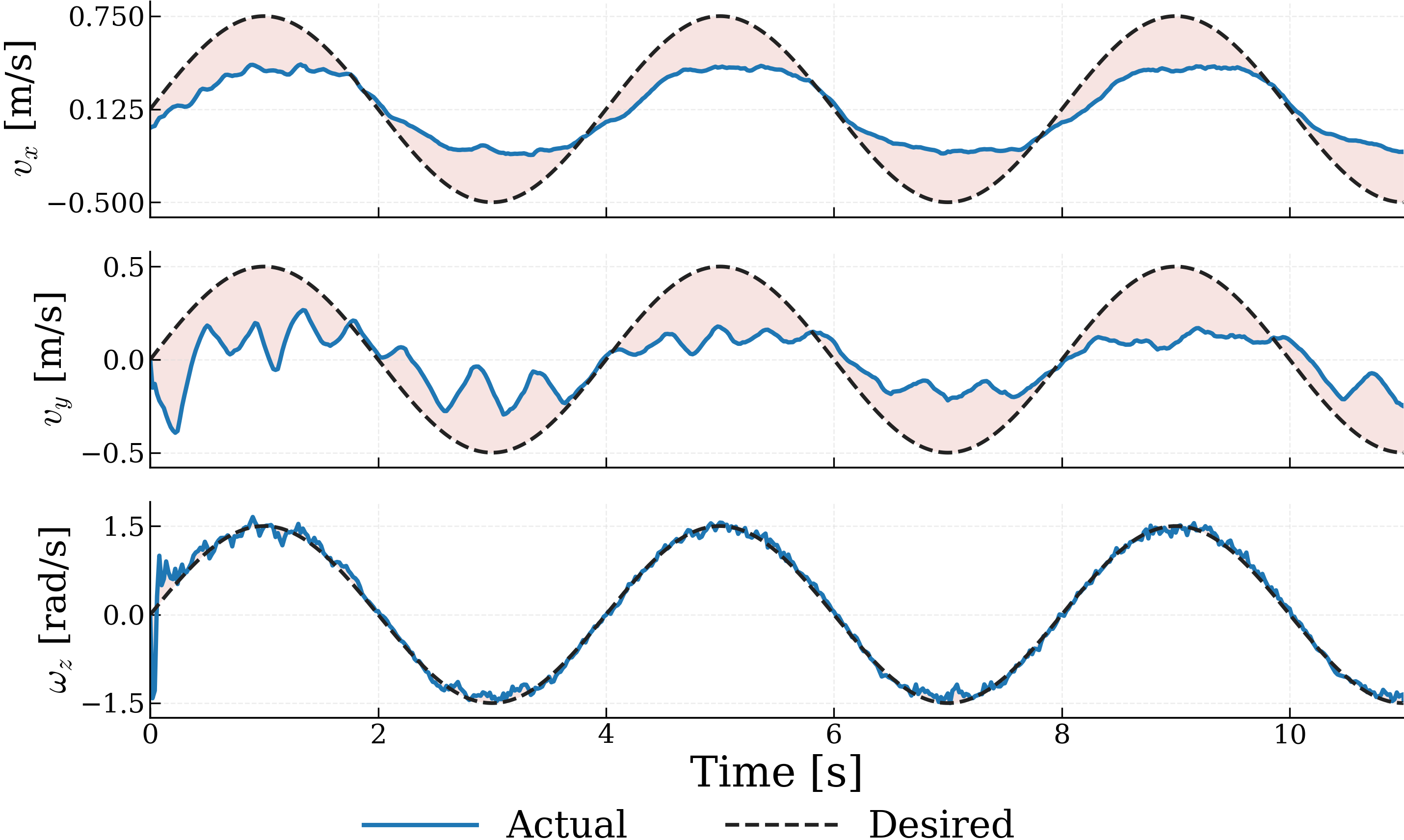}
    \caption{Policy velocity tracking performance averaged over $10$ trials. The Policy was tasked with following three sinusoidal trajectories, one for each of the $v_x$, $v_y$, and $\omega_z$ velocity commands.}
    \label{fig:velTrackPerf}
\end{figure}

Figure~\ref{fig:velTrackPerf} shows that the deployed Policy was able to follow the general sinusoidal command structure on hardware. The $v_x$ and $\omega_x$ commands were tracked with the clearest periodic response, while the lateral $v_y$ command showed larger oscillations and reduced tracking accuracy. Across the trials, the Policy tended to produce smoothed, attenuated responses rather than matching the commanded amplitude. This behavior is expected for a small humanoid platform, where aggressive velocity tracking can conflict with balance and contact stability.

\subsection{Multi-Terrain Locomotion Metrics}
On rough terrain, the Policy maintained general locomotion despite moderate surface variation. Larger vertical changes were more difficult, and the robot often slowed down, stepped away from, or avoided terrain regions that appeared likely to destabilize the hardware. In these cases, the Policy would sometimes ignore commanded velocity to preserve stability. This suggests that the local terrain observations helped the Policy learn a conservative safety behavior: when nearby terrain features indicated a ledge or dangerous height change, the controller prioritized avoiding termination over tracking the velocity command. While this improves hardware safety, it also limits traversal over more aggressive rough terrain.

\section{CONCLUSION}
This work presented a low-cost terrain-sensing approach for humanoid locomotion using a compact ToF height scan and a token-compressed temporal transformer policy. The method combines proprioceptive measurements, velocity commands, previous actions, and sparse local terrain samples through tokenization, self-attention, latent token compression, and temporal cross-attention over a rolling history. Hardware experiments on the Robotis OP3 showed that the Policy could track sinusoidal velocity commands and perform rough-terrain walking using the local terrain features. The main limitation observed was conservative behavior near larger vertical terrain changes, where the Policy often reduced or ignored commanded motion to preserve stability. Future work will focus on improving terrain traversal over larger height discontinuities, refining velocity tracking, and evaluating whether additional terrain coverage or richer local sensing improves robustness without increasing sensing cost.





\bibliographystyle{./format/IEEEtran.bst}
\bibliography{./format/IEEEabrv.bib,references}
\end{document}